# Poisson Kernel Avoiding Self-Smoothing in Graph Convolutional Networks

**Ziqing Yang**[1], **Shoudong Han**[1]* and **Jun Zhao**[2]
[1] National Key Laboratory of Science and Technology on Multispectral Information Processing,
School of Artificial Intelligence and Automation, Huazhong University of Science and Technology
[2] Nanyang Technological University
{ziqingyang, shoudonghan}@hust.edu.cn, junzhao@ntu.edu.sg

## Abstract

Graph convolutional network (GCN) is now an effective tool to deal with non-Euclidean data, such as social networks in social behavior analysis, molecular structure analysis in the field of chemistry, and skeleton-based action recognition. Graph convolutional kernel is one of the most significant factors in GCN to extract nodes' feature, and some improvements of it have reached promising performance theoretically and experimentally. However, there is limited research about how exactly different data types and graph structures influence the performance of these kernels. Most existing methods used an adaptive convolutional kernel to deal with a given graph structure, which still not reveals the internal reasons. In this paper, we started from theoretical analysis of the spectral graph and studied the properties of existing graph convolutional kernels. While taking some designed datasets with specific parameters into consideration, we revealed the self-smoothing phenomenon of convolutional kernels. After that, we proposed the Poisson kernel that can avoid self-smoothing without training any adaptive kernel. Experimental results demonstrate that our Poisson kernel not only works well on the benchmark dataset where state-of-the-art methods work fine, but also is evidently superior to them in synthetic datasets.

## 1 Introduction

The fundamental theory of graph convolutional neural network is spectral graph wavelet transform (SGWT) [Hammond *et al.*, 2011], which generalized the Fourier transform and wavelet transform of traditional signals to the spectral analysis of graphs. It was pointed out that the operation of wavelet transform in graph field is to define a wavelet transform operation function $T_g$ on Laplacian matrix $L$ of graph. After the use of polynomial approximation theory [Phillips, 2003], the approximation form was given by the Chebyshev polynomial function. The $k^{th}$ Chebyshev polynomial $T_k(x)$ is defined as:

$$T_{k+2}(x) = 2xT_{k+1}(x) - T_k(x), T_0(x) = 1, T_1(x) = x \quad (1)$$

On this basis, the most classical graph convolutional network (GCN) model [Kipf and Welling, 2017] was proposed, which was the first one to apply Chebyshev polynomial approximation form to Graph neural networks (GNNs). It was the pioneering work in this field. The layer-wise propagation formula of the classical semi-supervised classification was also defined as GCN model.

$$H^{(l+1)} = \sigma(\mathbb{F}(\mathcal{G})H^{(l)}W^{(l)}) \quad (2)$$

Here $H^{(l)} \in \mathbb{R}^{n \times d}$ is the feature matrix in the $l^{th}$ layer, $W^{(l)} \in \mathbb{R}^{d \times d}$ is a trainable weight matrix and $\sigma(\cdot)$ denotes a nonlinear activation function. Besides, $\mathbb{F}(\mathcal{G})$ is the graph convolutional kernel which is a function or matrix defined on the graph $\mathcal{G}$ and it is used to extract nodes' feature.

In GCN model, the form of $\mathbb{F}(\mathcal{G})$ can be obtained by simplifying the approximation form of Chebyshev polynomial to the first order approximation.

$$Z = \widehat{D}^{-\frac{1}{2}} \widehat{A} \widehat{D}^{-\frac{1}{2}} X \Theta \quad (3)$$

Here $\widehat{A}$ is the adjacency matrix of the graph and $\widehat{D}$ is the degree matrix.

Formula (3) defined the GNN convolution kernel based on graph adjacency matrix. After this work, many other improvements of convolution kernel were proposed, for example, SGC model [Wu *et al.*, 2019a] pointed out that $(\widehat{D}^{-\frac{1}{2}} \widehat{A} \widehat{D}^{-\frac{1}{2}})^k$ can include further neighbor information.

GNNs have been applied successfully in many tasks, especially for dealing with non-Euclidean data, such as attributed graph clustering [Zhang *et al.*, 2019], social behavior analysis [Wu *et al.*, 2019b; Wu *et al.*, 2018; Li *et al.*, 2019a], chemical and biological classification [Rhee, Sungmin et al., 2017; Daller *et al.*, 2018; Xu *et al.*, 2017; Ohue *et al.*, 2019], labeled images detection [Chen *et al.*, 2019], learning human-object interactions [Qi *et al.*, 2018] and skeleton-based action recognition [Yan *et al.*, 2018; Shi *et al.*, 2019; Li *et al.*, 2019b; Huang *et al.*, 2019]. There still remains a question, is there a best graph convolution kernel which can perform well on any data? Adaptive filter graph neural network (AFGNN) [Wang *et al.*, 2020] model answered that such a best kernel does not exist and they used an evaluation standard to evaluate the performance of different convolutional kernels, and then put forward the model AFGNN of adaptive convolutional kernel. AGCN [Li *et al.*, 2018a] model also proposed that some potential connections between nodes should be found out to satisfy more situations. But how different graph structures influence the performance of different kernels? Adaptive kernels only gave us a result and told us a kernel after training can adapt to a certain situation. There is limited research about how exactly different data types and graph structures influence the performance of these kernels, and if we find out the internal reasons, they must be a good guidance for us to design or train a better kernel.

---
* Corresponding Author

We summarize our contributions as follows:

1.The properties of graph convolutional kernels are studied and we reveal the self-smoothing phenomenon of convolutional kernels in order to find out how they work in some designed data structures.

2.We propose the Poisson kernel, which can work well on most situation without training any adaptive kernel.

## 2 Properties of Graph Convolutional Kernels

The layer-wise propagation formula of graph convolutional has been defined in formula (2). In this section, we will study the properties of graph convolutional kernel $\mathbb{F}(\mathcal{G})$ and find out how it works in some designed datasets.

Before our analysis begins, some notations and definitions are given as follows.

### 2.1 Preliminaries

$\mathcal{G} = (\mathcal{V}, E, X)$ represents an undirected graph defined with node set $\mathcal{V} = \{v_1, v_2, \ldots, v_n\}$ and edge set $E$, here $n = |\mathcal{G}|$ shows the number of nodes in $\mathcal{G}$ and we use $(v_i, v_j)$ to represent an edge between node $v_i$ and $v_j (i \neq j)$. $X = \{x_1, x_2, \ldots, x_n\} \in \mathbb{R}^{n \times d}$ is the $d$-dimensional feature matrix of all nodes and $x_i \in \mathbb{R}^d$. If we only need single feature, that is $d=1$, $X$ is a vector in $\mathbb{R}^n$.

$(a_{ij})_{m \times n}$ is the form of a matrix whose element on the $i^{th}$ row and $j^{th}$ column is $a_{ij}$. In the following text, we set $m = n = |\mathcal{G}|$ by default and use the form of $(a_{ij})$ for simplicity.

The value of elements in the adjacency matrix $A$ without self-circle is defined as $A_{ij}(i \neq j) = \begin{cases} 1 & (v_i, v_j) \in E \\ 0 & (v_i, v_j) \notin E \end{cases}$ and $A_{ii} = 0$. Degree matrix $D = diag(\sum_{j=1}^n A_{ij})$ is defined as a diagonal matrix of the degree of each node.

The matrix with self-circle is defined as $\hat{A} = A + I$ and $\widehat{D} = D + I = diag(d_i)$, here $d_i = \sum_{j=1}^n A_{ij} + 1 \geq 1$. The symmetric-normalized Laplacian is defined as $L = I - D^{-\frac{1}{2}} A D^{-\frac{1}{2}}$ and the eigenvalue of $L$ is $\lambda$.

The kernel proposed in formula (3) is defined as

$$\hat{L} = \widehat{D}^{-\frac{1}{2}} \hat{A} \widehat{D}^{-\frac{1}{2}} = \left(\frac{A_{ij}}{\sqrt{d_i d_j}}\right) \quad (4)$$

which is called Laplacian matrix of $\mathcal{G}$ and the eigenvalue of $\hat{L}$ is $\hat{\lambda}$.

### 2.2 The Structure and Parameters of Graph $\mathcal{G}$

An approach [Wang et al., 2020] to generate synthetic dataset based on stochastic block model [Holland et al., 1983] tells us the three key factors to build a graph dataset, which are nodes' label (or category) $Y$, the feature vector $X$ and the adjacency matrix $A$. In a certain label $Y$, we assume that the feature of a node is sampled from the same distribution determined by its corresponding label. Furthermore, the edges are generated via Bernoulli distributions independently and the parameters of Bernoulli distribution are determined by the classification of the corresponding pair nodes $v_i$ and $v_j$, that is $A_{ij}|Y_i, Y_j \sim Ber(p_{\{Y_i, Y_j\}})$. If $Y_i = Y_j$, we define $p_{\{Y_i, Y_j\}} = p$ and if $Y_i \neq Y_j$, $p_{\{Y_i, Y_j\}} = q$.

Actually, $p$ shows the probability of connecting two nodes with the same label and $q$ shows the opposite one. In real world, we can assume that $p>q$, which means the connection relationship between two individuals with the same label is stronger than that with the different labels.

We call $(p + q)/2 = \rho$ as the "Density" of a graph and $|p - q| = \varepsilon$ as the "Density Gap".

### 2.3 Self-Smoothing of Convolutional Kernels

SGC model [Wu et al., 2019a] shows that in graph convolution networks, higher power of $\hat{L}$ can obtain the feature from further neighbor. If we remove the nonlinear activation function in formula (2), we have

$$H^{(l+1)} = \hat{L}^{l+1} H^{(0)} \prod_{i=0}^l W^{(i)} = \hat{L}^{l+1} X W \quad (5)$$

This formula leads to a question: is it true that the larger $k$ value is, the further neighbor information can be included and the better for feature extraction? In fact, we have the following results and over-smoothing Theorem 1.

**Lemma 1**. $\hat{L}$ has an eigenvalue of 1, and the corresponding eigenvector is $u = \left(\sqrt{d_1} \; \sqrt{d_2} \; \cdots \; \sqrt{d_n}\right)^T$.

*Proof.* We only need to verify that $\hat{L}u = u$.

Notice that $d_i = \sum_{j=1}^n A_{ij}$, and $\hat{L} = \left(\frac{A_{ij}}{\sqrt{d_i d_j}}\right)$, so

$$\hat{L}\begin{pmatrix}\sqrt{d_1}\\\sqrt{d_2}\\\vdots\\\sqrt{d_n}\end{pmatrix} = \begin{pmatrix}\frac{1}{\sqrt{d_1}}\sum_{j=1}^n \frac{1}{\sqrt{d_j}} A_{1j}\sqrt{d_j}\\\frac{1}{\sqrt{d_2}}\sum_{j=1}^n \frac{1}{\sqrt{d_j}} A_{2j}\sqrt{d_j}\\\vdots\\\frac{1}{\sqrt{d_n}}\sum_{j=1}^n \frac{1}{\sqrt{d_j}} A_{nj}\sqrt{d_j}\end{pmatrix} = \begin{pmatrix}\sqrt{d_1}\\\sqrt{d_2}\\\vdots\\\sqrt{d_n}\end{pmatrix}$$

□

**Lemma 2.** $-1 < \hat{\lambda}_1 \leq \hat{\lambda}_2 \leq \cdots \leq \hat{\lambda}_n = 1$, here the first inequality is guaranteed to be strictly less than.

*Proof.* If a graph has no bipartite components, all eigenvalues of $I - \hat{L}$ fall in [0,2) [Chung, 1997], so that all eigenvalues of $\hat{L}$ will fall in (-1,1]. □

**Theorem 1** (Over-smoothing of Laplacian). *If $\mathcal{G}$ has only one connected component, $\hat{L} = \widehat{D}^{-\frac{1}{2}} \hat{A} \widehat{D}^{-\frac{1}{2}} = U diag(\hat{\lambda}_i) U^T$ is the symmetric-normalized Laplacian and $U =*

$(u_1, u_2, \ldots u_n)$ is the orthogonal normalized eigenvector, we have:
$$\lim_{k \to \infty} \hat{L}^k = U \begin{pmatrix} 0 & & & \\ & \ddots & & \\ & & 0 & \\ & & & 1 \end{pmatrix} U^T = \left( \frac{\sqrt{d_i d_j}}{\sum_{i=1}^n d_i} \right)$$
where $u_n = \frac{1}{\sqrt{\sum_{i=1}^n d_i}} \left( \sqrt{d_1} \ \sqrt{d_2} \ \cdots \ \sqrt{d_n} \right)^T$ is the eigenvector of 1 which satisfies $\hat{L} u_n = u_n$.

*Proof.* We have
$$\hat{L}^k = U diag(\hat{\lambda}_i^k) U^T$$
By Lemma 2, we have:
$$\lim_{k \to \infty} \hat{L}^k = U diag\left( \lim_{k \to \infty} \hat{\lambda}_i^k \right) U^T$$
$$= U \begin{pmatrix} 0 & & & \\ & \ddots & & \\ & & 0 & \\ & & & 1 \end{pmatrix} U^T = (u_{ni} u_{nj})$$
$u_n = (u_{n1} \ u_{n2} \ \cdots \ u_{nn})^T$ is the orthogonal-normalized eigenvector of $\hat{\lambda}_n = 1$. By Lemma 1, we normalize the vector $\left( \sqrt{d_1} \ \sqrt{d_2} \ \cdots \ \sqrt{d_n} \right)^T$ to
$$u_n = \frac{1}{\sqrt{\sum_{i=1}^n d_i}} \left( \sqrt{d_1} \ \sqrt{d_2} \ \cdots \ \sqrt{d_n} \right)^T$$
Then we get
$$\lim_{k \to \infty} \hat{L}^k = (u_{ni} u_{nj}) = \frac{1}{\sum_{i=1}^n d_i} \left( \sqrt{d_i d_j} \right) \qquad \square$$

Here we define
$$S = \lim_{k \to \infty} \hat{L}^k = \left( \frac{\sqrt{d_i d_j}}{\sum_{i=1}^n d_i} \right) \tag{6}$$

In fact, if graph $\mathcal{G}$ has multiple connected components, its adjacency matrix is the block-diagonal form of the adjacency matrix defined in each connected subgraph. The general result can be concluded in some other work [Li *et al.*, 2018b]. For the sake of narrative, we assume that the graphs in our question have only one connected component.

Now let us take $S$ into account. According to Theorem 1, it is easy to find that $S$ is idempotent, and if we use it to deal with $X = \{x_1, x_2, \ldots, x_n\} \in \mathbb{R}^n$, we have
$$SX = S^k X = \frac{\sum_{j=1}^n \sqrt{d_j} x_j}{\sum_{i=1}^n d_i} \begin{pmatrix} \sqrt{d_1} \\ \sqrt{d_2} \\ \vdots \\ \sqrt{d_n} \end{pmatrix} \tag{7}$$

If we use $S$ as the convolution kernel, no matter how many linear transformations we take, the extracted features will always be parallel to the vector $\left( \sqrt{d_1} \ \sqrt{d_2} \ \cdots \ \sqrt{d_n} \right)^T$. It is easy to see that this phenomenon is a corollary of the following Theorem 2 of self-smoothing.

**Theorem 2** (Self-smoothing of graph convolutional kernels). *If $\mathbb{F}(\mathcal{G})$ is idempotent, for any vector $X^{(0)} \in \mathbb{R}^n$, any mapping iteration $\tilde{X}^{(l)} = \mathbb{F}(\mathcal{G}) X^{(l)} \omega^{(l)}$, let $\mathcal{V}_i$ be a subspace of $\mathbb{R}^n$ that can contain all $\tilde{X}^{(l)}$, then*
$$\min_i \dim \mathcal{V}_i = tr(\mathbb{F}(\mathcal{G})) = rank(\mathbb{F}(\mathcal{G})) \leq n$$

*Proof.* Let $u_1, u_2, \cdots u_n$ be the orthonormal eigenvectors of $\mathbb{F}(\mathcal{G})$'s eigenvalues $\lambda_1, \lambda_2, \cdots \lambda_n$, $rank(\mathbb{F}(\mathcal{G})) = r$. $\mathbb{F}(\mathcal{G})$ is an idempotent matrix so $\lambda_1 = \cdots = \lambda_r = 1, \lambda_{r+1} = \cdots = \lambda_n = 0$. For any vector $X^{(l)} \in \mathbb{R}^n$, consider $\mathbb{R}^n$ spanned by $u_1, u_2, \cdots u_n$, let $X^{(l)} = \sum_{i=1}^n x_i u_i$, we have:
$$\tilde{X}^{(l)} = \mathbb{F}(\mathcal{G}) X^{(l)} \cdot \omega^{(l)} = \mathbb{F}(\mathcal{G}) \sum_{i=1}^n x_i u_i \cdot \omega^{(l)}$$
$$= \sum_{i=1}^n x_i \mathbb{F}(\mathcal{G}) u_i \cdot \omega^{(l)} = \sum_{i=1}^r x_i u_i \cdot \omega^{(l)}$$
The linear space
$$\mathcal{V} = \left\{ \sum_{i=1}^r \alpha_i u_i \ | \alpha_1, \alpha_2, \cdots \alpha_r \in \mathbb{R} \right\}$$
is a subspace of $\mathbb{R}^n$ that can contain all $\tilde{X}^{(l)}$, and it is obvious that any other subspace of $\mathbb{R}^n$ with less dimension than $\mathcal{V}$ can not contain all $\tilde{X}^{(l)}$. And $tr(\mathbb{F}(\mathcal{G})) = \sum_{i=1}^n \lambda_i = r$, so
$$\dim \mathcal{V} = r = tr(\mathbb{F}(\mathcal{G})) \leq n \qquad \square$$

Theorem 2 tells us that if we use an idempotent matrix as the convolution kernel, our feature mapping is going to be always restricted into a fixed lower dimensional subspace.

If $\mathbb{F}(\mathcal{G})$ is idempotent, we call it a self-smoothing kernel. Obviously, $S$ is self-smoothing.

In fact, we know that the linear transformation corresponding to an idempotent matrix represents the projection transformation in linear space. As shown in Figure 1, the projection direction of an idempotent matrix $A$ is the corresponding eigenvector of eigenvalue 1.

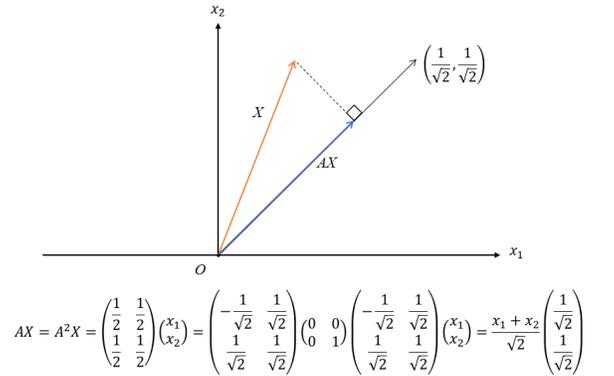

Figure 1: Linear transformation of an idempotent matrix

### 2.4 Convolutional Kernels in Specific Graphs

In this section, we are going to reveal how convolutional kernels work in specific graph structures. The parameters of these graphs are well designed.

**Graph with SmallGap**

A graph with SmallGap means the parameter $|p - q| = \varepsilon$ is small. In this structure, we consider the two-category scenario. Let $n_1, n_2$ be the node amount of two categories and $n_1 + n_2 = N$, density $(p + q)/2 = \rho$, density gap $|p - q| = \varepsilon$, label ratio $n_1/n_2 = r$.

The expectation of adjacency matrix is

$$\hat{A} = \begin{pmatrix} 1 & \cdots & p & & & \\ \vdots & \ddots & \vdots & & (q)_{n_1 \times n_2} & \\ p & \cdots & 1 & & & \\ & & & 1 & \cdots & p \\ & (q)_{n_2 \times n_1} & & \vdots & \ddots & \vdots \\ & & & p & \cdots & 1 \end{pmatrix}$$

and $d_1 = 1 + (n_1 - 1)p + n_2 q$, $d_2 = 1 + (n_2 - 1)p + n_1 q$ are the degree of nodes in two categories.

We directly calculate the eigenvalues of $\hat{L} = \hat{D}^{-\frac{1}{2}} \hat{A} \hat{D}^{-\frac{1}{2}}$ by mathematical induction and get

$$\hat{\lambda} = \begin{cases} (1 - \rho)/d_1 & (\text{multiplicity}: (n_1 - 1)) \\ (1 - \rho)/d_2 & (\text{multiplicity}: (n_2 - 1)) \\ 1 - (n_2/d_1 + n_1/d_2)q \\ 1 \end{cases}$$

In a graph with SmallGap, assume that $\varepsilon \to 0$, then
$d_1$ and $d_2 \to 1 + (N - 1)\rho$, $p$ and $q \to \rho$

So

$$\hat{\lambda} \to \begin{cases} \frac{1-\rho}{1+(N-1)\rho} & (\text{multiplicity}: (N - 1)) \\ 1 \end{cases} \quad (8)$$

If there are many nodes in graph $\mathcal{G}$, that is $N$ is sufficiently large, then the eigenvalues of $\hat{L}$ is sufficiently close to zero according to formula (8) and it will lead to self-smoothing like $S$. That means if the connection of nodes in the same category and the different categories are slightly different, use $\hat{L}$ to do classification is more difficult.

If $\rho=1$, $\varepsilon=0$ extremely, that is $p=q=1$, we have:

$$\hat{\lambda} = \begin{cases} 0 & (\text{multiplicity}: (N - 1)) \\ 1 \end{cases}$$

In fact, the graph structure at this time is a fully connected graph, and there are edges between any pair of nodes in the same classification or not. Now $\hat{L} = (1/n)$ itself is self-smoothing. If we use $\hat{L}$ directly as the convolution kernel to handle the classification, according to Theorem 2, the result must be inappropriate.

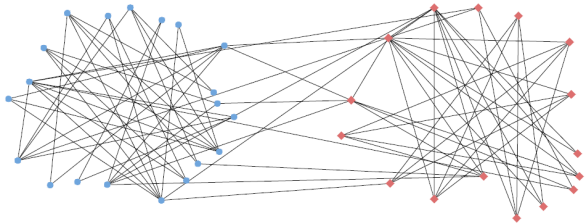

Subgraph 1 with SmallGap    Subgraph 2 with SmallGap

Figure 2: A graph with LargeGap, the subgraph inside each category is SmallGap.

**Graph with LargeGap**

A graph with LargeGap refers to those with large parameter $|p - q| = \varepsilon$. Consider the simplest binary classification problems, we consider an extreme situation where $q$ is sufficiently close to zero, then the graph is clearly separated into two connected subgraphs according to the category. It is easy to do classification only through graph's structure. But inside the subgraph, its structure is completely consistent with a graph with SmallGap. As shown in Figure 2, a graph with Large Gap is a combination of two subgraph with SmallGap. Inside each subgraph, if we want to continue classification in this category and divide nodes into smaller categories, surely using $\hat{L}$ as the convolution kernel is not suitable. According to the above analysis we can get the following conclusion, if we use $\hat{L}$ as convolution kernel, it is easier to categorize closely connected nodes into one category, and more difficult to continue to categorize these closely connected nodes.

### 2.5 Summaries

From the discussion in Section 2.4, we can get the following summaries about the convolution kernel $\mathbb{F}(\mathcal{G})$:

1. Whatever the graph structure, $\mathbb{F}(\mathcal{G})$ must have a fixed eigenvalue which tells the connected components of a graph, such as $\hat{L} = \hat{D}^{-\frac{1}{2}} \hat{A} \hat{D}^{-\frac{1}{2}}$ will always have an eigenvalue of 1.

2. The more eigenvalues of 1 $\mathbb{F}(\mathcal{G})$ have, the more original features of all nodes in the graph are retained, and the deeper features cannot be extracted.

3. The more eigenvalues of 0 $\mathbb{F}(\mathcal{G})$ have, the easier it is for $\mathbb{F}(\mathcal{G})$ to achieve self-smoothing results in feature mapping processing. According to Theorem 1, the network will soon converge. When we are designing new convolution kernels, we should try to avoid those matrices with too many eigenvalues of 0, that is, for any graph structure, $\mathbb{F}(\mathcal{G})$ cannot be self-smoothing.

## 3. Our Graph Convolution Kernel

In this paper, the form of graph convolution kernel designed for avoiding self-smoothing in the whole graph domain is:

$$P = (1 - r^2)\left((r^2 + 1)I - 2r\hat{L}\right)^{-1} \quad (9)$$

We call it Poisson Kernel.

### 3.1 Eigenvalue Mapping

The first order Chebyshev polynomial approximation form of graph convolutional kernel is defined in formula (3). Derivation process of it is generalized from the Chebyshev polynomial approximation form [Hammond *et al.*, 2011]

$$H^{(l+1)} = \sigma\left(\sum_{k=0}^{K} \theta'_k T_k(\tilde{L}) H^{(l)}\right) \quad (10)$$

Here $\tilde{L} = \frac{2}{\lambda_{max}} L - I$, $\lambda_{max}$ is the largest eigenvalue of $L$, and $T_k(x)$ is the $k^{th}$ Chebyshev polynomial defined in formula (1).

We let $\lambda_{max} = 2$, $K=1$ and $\theta'_0 = -\theta'_1 = \omega$, we have:

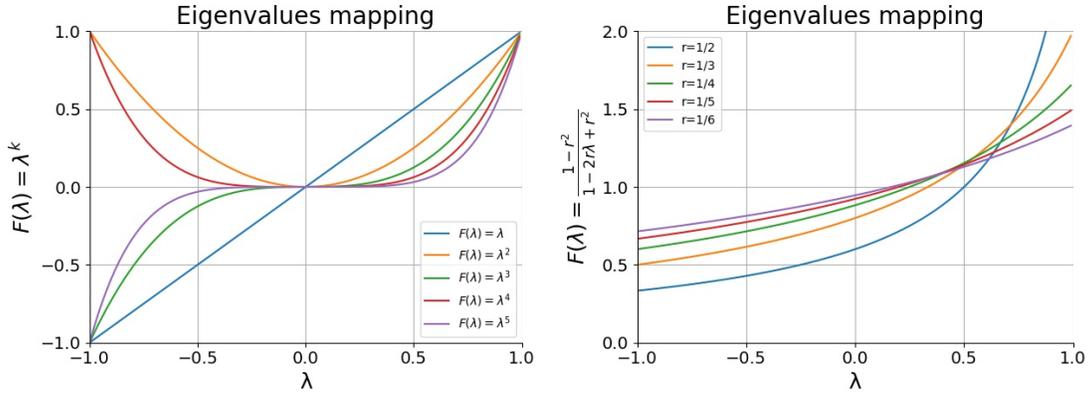

Figure 3: Eigenvalue mapping of $\lambda^k$ and Poisson kernel $\frac{1-r^2}{1-2r\lambda+r^2}$

$$\sum_{k=0}^{K} \theta'_k T_k(\tilde{L}) \approx \omega \left(I + D^{-\frac{1}{2}} A D^{-\frac{1}{2}}\right) \quad (11)$$

After a re-normalization trick:

$$I + D^{-\frac{1}{2}} A D^{-\frac{1}{2}} \to \hat{D}^{-\frac{1}{2}} \hat{A} \hat{D}^{-\frac{1}{2}} \quad (12)$$

We have the final form of convolution kernel in formula (4) which is $\mathbb{F}(\mathcal{G}) = \hat{L} = \hat{D}^{-\frac{1}{2}} \hat{A} \hat{D}^{-\frac{1}{2}}$.

In SGC [Wu *et al.*, 2019a],

$$\mathbb{F}(\mathcal{G}) = \hat{L}^k \quad (13)$$

Even there are many different forms of convolutional kernels, they are all an eigenvalue mapping $\mathbb{F}(\hat{\lambda})$ of $\hat{L}$, in SGC $\mathbb{F}(\hat{\lambda}) = \hat{\lambda}^k$. In formula (9), we actually built a mapping of $\hat{\lambda}$:

$$\mathbb{F}(\hat{\lambda}) = \frac{1-r^2}{1-2r\hat{\lambda}+r^2} \quad (14)$$

This formula can map $\hat{\lambda} \in (-1,1]$ into $\left(\frac{1-r}{1+r}, \frac{1+r}{1-r}\right]$. After mapping, the kernel $\mathbb{F}(\mathcal{G})$ will not contain any eigenvalue of zero, which will avoid self-smoothing of $\mathbb{F}(\mathcal{G})$.

In fact, we can build other mapping to avoid self-smoothing of $\mathbb{F}(\mathcal{G})$, such as linear mapping trick and the corresponding $\mathbb{F}(\mathcal{G})$ is defined as

$$\mathbb{F}(\mathcal{G}) = Li = \frac{1}{2}(I + \hat{L}) \quad (15)$$

whose form is similar to the re-normalization trick in formula (12). Different mappings are shown in Figure 3. Our Poisson kernel will not contain any eigenvalue of zero which will lead to self-smoothing.

### 3.2 Proposed Poisson Kernel

We can see $\hat{L} = \hat{D}^{-\frac{1}{2}} \hat{A} \hat{D}^{-\frac{1}{2}}$ only contains the first-order neighbor characteristics of Chebyshev polynomial, and Laplacian of high power can contain further. We do the following treatment to Chebyshev polynomial coefficients, let $\theta'_0 = 1, \theta'_k = 2r^k (k \geq 1), r \in (-1,1)$ in formula (10), we have:

$$\mathbb{F}(\mathcal{G}) = \sum_{k=0}^{\infty} \theta'_k T_k(\tilde{L}) = 1 + 2 \sum_{k=1}^{\infty} r^k T_k(\tilde{L})$$

Compared with formula (11), this form can contain higher order Chebyshev polynomials.

Actually,

$$1 + 2\sum_{k=1}^{\infty} r^k T_k(\tilde{L}) = (1-r^2)\left((r^2+1)I - 2r\tilde{L}\right)^{-1} \quad (16)$$

Before we prove this fact, we have to prove the following two lemmas. Here we set a matrix $L$ with all eigenvalues in range [-1,1] and a real number $r \in (-1,1)$.

**Lemma 3**. $\left((r^2+1)I - 2rL\right)$ *is invertible.*

*Proof.* We only need to prove that all eigenvalues of $\left((r^2+1)I - 2rL\right)$ are not zero. Let $\lambda$ be the eigenvalue of $L$ and $\lambda'$ be the eigenvalue of $\left((r^2+1)I - 2rL\right)$, and

$$\lambda' = (r^2+1) - 2r\lambda = (r-\lambda)^2 + 1 - \lambda^2$$

Obviously $\lambda' = 0$ if and only if $r = \lambda = \pm 1$, and we set $r \in (-1,1)$, so $\lambda' \neq 0$. □

**Lemma 4**. $\lim_{K \to \infty} r^K T_K(L) = 0$

*Proof.* Let $\lambda_i$ be the eigenvalues of $L$ and we can set $\lambda_i = \cos(\theta_i)$ for $\lambda_i \in [-1,1]$. According to the property of Chebyshev polynomials we have

$$r^K T_K(L) = r^K U T_K(diag(\lambda_i)) U^T$$
$$= r^K U diag(T_K(\lambda_i)) U^T = r^K U diag(cosK\theta_i) U^T$$

$cosK\theta_i$ is bounded, $\lim_{K \to \infty} r^K T_K(L) = 0$ for $|r| < 1$. □

By Lemma 3 and 4, we have Theorem 3.

**Theorem 3.**

$$\sum_{k=0}^{\infty} r^k T_k(L) = \left((r^2+1)I - 2rL\right)^{-1}(I - rL)$$

*Proof.* We let

| Model & kernel | GCN | | | | SGC | | | | |
|---|---|---|---|---|---|---|---|---|---|
| | $\hat{L}$ | $S$ | $Li$ | **P(Ours)** | $\hat{L}$ | $\hat{L}^2$ | $S$ | $Li$ | **P(Ours)** |
| Cora | 80.78 | 18.55 | 77.61 | **81.35** | 81.17 | **81.60** | 18.74 | 77.15 | **80.66** |
| Citeseer | **71.21** | 19.54 | 70.28 | **68.97** | 72.40 | **72.60** | 20.50 | 71.80 | **71.00** |
| SmallGap | 83.50 | 47.88 | 99.96 | **99.98** | 74.70 | 48.30 | 48.30 | 99.95 | **99.96** |
| SmallRatio | 83.00 | 80.00 | 87.89 | **88.15** | 87.29 | 81.17 | 80.00 | 84.80 | **83.26** |

Table 1. Testing accuracy of different kernels on GCN and SGC

$$P = \sum_{k=0}^{K} r^k T_k(L) = I + rL + \sum_{k=2}^{K} r^k T_k(L)$$

We can also get

$$-2rLP = -2rL - \sum_{k=2}^{K} 2r^k LT_{k-1}(L) - 2r^{K+1}LT_K(L)$$

and

$$r^2 P = \sum_{k=2}^{K} r^k T_{k-2}(L) + r^{K+1}T_{K-1}(L) + r^{K+2}T_K(L)$$

Sum these three equations up, noticing that $T_k(L) - 2LT_{k-1}(L) + T_{k-2}(L) = 0$, we have
$((r^2 + 1)I - 2rL)P = I - rL - r^{K+1}T_{K+1}(L) + r^{K+2}T_K(L)$

Let $K \to \infty$, by Lemma 3 and Lemma 4 we get our Theorem 3 proved. □

By Theorem 3 we have

$$\mathbb{F}(\mathcal{G}) = 1 + 2\sum_{k=1}^{\infty} r^k T_k(\tilde{L})$$

$$= 2\sum_{k=0}^{\infty} r^k T_k(\tilde{L}) - 1 = (1 - r^2)\left((r^2 + 1)I - 2r\tilde{L}\right)^{-1}$$

This is formula (16).

$\tilde{L} = -D^{-\frac{1}{2}}AD^{-\frac{1}{2}}$ and $\hat{L} = \hat{D}^{-\frac{1}{2}}\hat{A}\hat{D}^{-\frac{1}{2}}$ have only self-circle difference on structure, we can replace $\tilde{L}$ as $\hat{L}$:

$$\mathbb{F}(\mathcal{G}) = P = (1 - r^2)\left((r^2 + 1)I - 2r\hat{L}\right)^{-1}$$

This form is similar to the Poisson Kernel in harmonic analysis:

$$P(r, \theta) = \frac{1 - r^2}{1 - 2r\cos\theta + r^2}$$
$$= 1 + 2\sum_{n=1}^{\infty} r^n \cos n\theta = \sum_{n=-\infty}^{+\infty} r^{|n|}e^{in\theta}$$

We call it Poisson Kernel in graph convolution which is defined in formula (9).

## 4 Experimental Results

We test the accuracy of some different kernels on GCN [Kipf and Welling, 2017] and SGC [Wu *et al.*, 2019a]. We only compare to kernels with the fixed form rather than adaptive ones, which need to be trained according to the structure of a given graph. If switching a new graph or dataset, you have to train the kernel again. The results are shown in Table 1.

In Table 1, $\hat{L}$, $\hat{L}^2$, $S$, $Li$ and $P$ are defined in formula (4), (13), (6), (15) and (9) respectively. We let $r = 0.5$ in our Poisson kernel $P$ without loss of generality.

SmallGap and SmallRatio are two synthetic datasets generated by SBM model [Holland *et al.*, 1983], the parameters of which can be found in [Wang *et al.*, 2020]. Generally speaking, $|p - q| = \varepsilon$ is small in SmallGap and label ratio $n_1/n_2 = r$ is small in SmallRatio.

From Table 1, it is easy to find that the accuracy of different models and kernels on benchmark datasets Cora and Citeseer are nearly identical but with slight differences. Our Poisson kernel still works well on these benchmark datasets where state-of-the-art methods work fine.

In the two synthetic datasets, especially in SmallGap, the accuracy of Poisson kernel is almost reaching 100%, which is superior to state-of-the-art methods. The classical kernels will lead to self-smoothing on SmallGap according to our former analysis. Compared with the adaptive kernels, our method does not need to train any adaptive kernel.

It can also be seen from Table 1 that GNNs with $S$ as the convolutional kernel work extremely worse than any others on every dataset, which is the inevitable consequence of our self-smoothing Theorem 2.

For the linear mapping trick $Li$ in formula (16) used to avoid self-smoothing, it still works very well on SmallGap, but it does not work well like Poisson kernel on benchmark datasets. We can draw the conclusion that Poisson kernel is a better kernel avoiding self-smoothing.

## 5 Conclusion

In this paper, we reveal the self-smoothing phenomenon in GCNs, which will reduce the performance of a graph convolutional kernel. In some special graph structures, the existing kernels will lead to self-smoothing inevitably. Our proposed Poisson kernel avoids this problem skillfully. Simultaneously, compared with other ways to avoid this problem, Poisson kernel still works well on the benchmark dataset where state-of-the-art methods work fine. Furthermore, our method doesn't have to train any adaptive kernel while changing data and it can be applied more generally in other scenarios.